\documentclass[manuscript,screen]{acmart}
\usepackage{caption}
\usepackage{subcaption}
\AtBeginDocument{%
  \providecommand\BibTeX{{%
    \normalfont B\kern-0.5em{\scshape i\kern-0.25em b}\kern-0.8em\TeX}}}

\setcopyright{acmcopyright}
\copyrightyear{2022}
\acmYear{2022}
\acmDOI{}

\acmConference[THRI]{Transactions on Human-Robot Interaction}
\acmBooktitle{Transactions on Human-Robot Interaction}

\begin{document}

\title[Human-Robot Joint Action Framework using AR and Eye Gaze]{Design and Implementation of a Human-Robot Joint Action Framework using Augmented Reality and Eye Gaze}

\author{Wesley P. Chan}
\affiliation{%
  \institution{Monash University}
  \country{Australia}
}

\author{Morgan Crouch}
\affiliation{%
  \institution{Monash University}
  \country{Australia}
}

\author{Khoa Hoang}
\affiliation{%
  \institution{Monash University}
  \country{Australia}
}

\author{Charlie Chen}
\affiliation{%
  \institution{Monash University}
  \country{Australia}
}

\author{Nicole Robinson}
\affiliation{%
  \institution{Monash University}
  \country{Australia}
}

\author{Elizabeth Croft}
\affiliation{%
  \institution{University of Victoria}
  \country{Australia}
}

\renewcommand{\shortauthors}{Chan et al.}

\begin{abstract}
When humans work together to complete a joint task, each person builds an internal model of the situation and how it will evolve. Efficient collaboration is dependent on how these individual models overlap to form a shared mental model among team members, which is important for collaborative processes in human-robot teams. The development and maintenance of an accurate shared mental model requires bidirectional communication of individual intent and the ability to interpret the intent of other team members. To enable effective human-robot collaboration, this paper presents a design and implementation of a novel joint action framework in human-robot team collaboration, utilizing augmented reality (AR) technology and user eye gaze to enable bidirectional communication of intent. We tested our new framework through a user study with 37 participants, and found that our system improves task efficiency, trust, as well as task fluency. Therefore, using AR and eye gaze to enable  bidirectional communication is a promising mean to improve core components that influence collaboration between humans and robots.

\end{abstract}



\maketitle

\section{INTRODUCTION}
        \label{sec:introduction}
        A key focus of human-robot interaction (HRI) research is to create seamless collaboration between robots and humans. Studies in psychological science indicate that a critical factor for collaboration between people, particularly when completing joint activities, is the capacity to reason about perception and action plans, and to construct a shared mental model of the collaborative task \cite{barron_achieving_2000}. This construction of a shared mental model is called ‘Theory of Mind’ \cite{baron-cohen_does_1985}. A shared mental model can be defined as the overlapping information that collaborating agents (where an agent can be a human or a robot) share about the task \cite{klimoski_team_1994}. The model is developed by first understanding each other's task representation (mutual understanding) and then incorporating both task representations (mutual agreement) \cite{dillenbourg_sharing_2006}. Once established, each agent then formulates their action plan to contribute to the task, and together, the team can proceed with joint actions to complete the task. The shared mental model allows for anticipatory information sharing, which can improve task efficiency \cite{van_den_bossche_team_2011}. 


\begin{figure}[t!]
    \centering
    \includegraphics[width=0.48\textwidth]{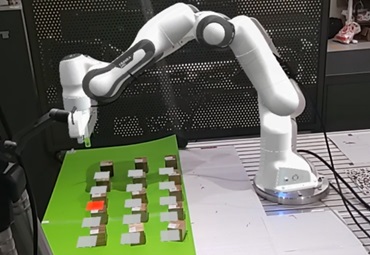}
    \caption{User's view when collaborating with the robot arm using our AR joint action framework. The block selected by the robot is highlighted in red to convey to the user the robot's intention of picking up the block.}
    \label{fig:overview}
\end{figure}


The establishment of an accurate shared mental model requires bidirectional communication between collaborating agents. Enabling effective communication within human-robot teams is a current challenge for the field of HRI, as humans and robots typically rely on different communicate channels to exchange information.
In recent years, augmented reality (AR) technology has grown as an interface technology, with a number of AR head-mounted displays becoming commercially available \cite{MagicLeap, kress201711, Moverio}. Leveraging AR technology, existing research have demonstrated its potential in proving intuitive communication between human and robot \cite{de_pace_systematic_2020,makris_augmented_2016,blankemeyer_intuitive_2018,quintero_robot_2018}
To enable effective human-robot teaming, this paper presents a novel human-robot joint action framework design incorporating AR technology. Our contributions are as follows: 
\begin{enumerate}
    \item We have proposed and implemented a novel design of a human-robot joint action framework, that uses AR and user eye gaze to enable bidirectional communication (Fig. \ref{fig:overview}).
    \item We conducted a user study with 37 participants to validate our novel system. Results show that our system improves task error and user's trust of the robot, without increasing the task load.
\end{enumerate}

While there exist research on the use of AR or eye gaze alone for human-robot collaboration, the combined use for implementing a joint action framework has not been proposed or examined.

\section{RELATED WORK}
        \label{sec:related_works}
        \subsection{Joint Action Frameworks}

A joint action framework should facilitate the coordination of individual agents' goals, intentions, plans, and actions during execution, towards establishing a shared mental models. In developing this shared model, Clodic et al. suggests that a framework must overcome motivational uncertainty (misaligned goals), instrumental uncertainty (shared plans and goals), and common ground (public information) uncertainty \cite{Clodic_Key_2017}. This is particularly important in human-robot joint action tasks as robots and humans (particularly non robot expert users) do not typically possess the same background knowledge or share the same level of mutual understanding as humans (particularly those who have shared experience) have with each other. 

Devin et al. has developed a joint action framework by proposing a Theory of Mind manager, which models an agent as a collection of sets. These sets include the high level actions that the agent is able to perform, the state of a goal, the set of facts about the environment, and the state of the plan from an agent's point of view. How the robot perceives the evolution of these states will determine how the robot will utilise its capacities to execute the shared plan. \cite{devin_implemented_2016}.
Breazeal et al. tried to reduce uncertainty and coordinate actions by replicating human-human collaborative discourse, such as recreating human social gestures, collaborative dialogue, and facial expressions in humanoids \cite{breazeal_teaching_2004}. These methods have shown to improve team outcomes but require a robot to have a level of anthropomorphism.
Lasota et al. showed that the speed, distance, and positioning of a robot arm can increase predictability and user confidence when working in close proximity \cite{lasota_toward_2014}. 
Nikolaidis et al. have also explored the used of unsupervised learning for improving joint action frameworks. Their system was able to learn a reward policy that can select a set of action sequences that outperform a manually specified policy in both efficiency and fluency of a human-robot team, by considering anticipatory actions \cite{Nikolaidis_efficient_2015}. 

\subsection{Augmented Reality in HRI}

Previous research on how AR can assist human-robot collaboration predominately focuses on using AR to display workplace information, control feedback information or task information \cite{de_pace_systematic_2020}. 
Examples of workspace information display uses AR to highlight the space that the robot is currently using or plans to use and where it is safe for the user. Such include using colour gradients projected onto a surface \cite{makris_augmented_2016} or using virtual light barriers placed within the environment \cite{vogel_novel_2017} to warn the user of the robot’s current operation.

Examples of control feedback information display include applications in path generation or teleoperation. Yamamoto et al. used AR 3D material property overlays and virtual fixtures to provide visual and haptic feedback in robotic surgery \cite{yamamoto_augmented_2012}. Blankemeyer et al. showed that AR virtual path displays can provide a more intuitive control mechanism \cite{blankemeyer_intuitive_2018}. Quintero et al. have also created an AR robot teleoperation interface, and showed that it reduces physical workload when compared with kinesthetic teaching, but at the expense of a higher mental workload \cite{quintero_robot_2018}.

Using Augmented Reality as a tool for displaying task and robot status information has been shown to improve HRI. Weng et al. explored and measured the effectiveness of AR displays including 3D arrows, virtual leading lines, and virtual text 
\cite{weng_robot_2019}.  
Walker et al. explored the use of AR for displaying an aerial robot’s future flight motion by visualizing navigation points, arrows, robot’s gaze, and an environment map. 
\cite{walker_communicating_2018, walker_robot_2019}. 

\subsection{Eye Gaze in HRI}

Eye gaze has been widely examined for HRI as it is a natural form of human communication that contains a wealth of information about one’s mental state. Eye gaze is used to ascertain what a person is looking at and to cue people's attention \cite{kuhn_eye_2010, frischen_gaze_2007}. 
Aronson et al. examined the dynamics of control behaviour and eye gaze in human-robot shared manipulation tasks. By studying gaze features like saccades, fixations, smooth pursuits, and scan paths, they determined that there were specific gaze patterns for monitoring and planning glances \cite{aronson_eye-hand_2018}. These provide meaningful signals that a robot could use to improve the quality of interaction. The use of these gaze patterns to predict intent was explored by Huang et al., who quantified patterns of gaze cues to create a support vector machine (SVM) based model. Their model had 76\% accuracy at predicting intended user requests using eye gaze alone. 
\cite{huang_using_2015}. 
They subsequently created an anticipatory control method for robots working in a team setting with humans. Using an intent prediction model for eye gaze, the robotic assistant was able to reach target objects 2.5 seconds earlier than a comparable reactive method. Additionally, user perception of the robot was found to be more positive when working with the robot incorporating the eye gaze model \cite{huang_anticipatory_2016}. They pointed out that the accuracy of theirs and other existing gaze prediction system \cite{mutlu_nonverbal_2009} is a limiting factor. 

 \section{RESEARCH AIM}
        \label{sec:research_aim}

While there exists research that use human eye gaze to enable a robot to predict the user's intent, and research that uses AR to display robot status and plans to the user, prior work that utilizes both AR and user eye gaze in the context of a joint action framework has not been reported.
Our research aims to construct a joint action framework that utilizes augmented reality and human eye gaze cues to enable bidirectional communication for human-robot collaboration, and investigate the benefits this approach can bring. 

We hypothesise that such a joint action framework for supporting human-robot collaborative tasks will improve task efficiency and fluency, reduce the user’s cognitive load and increase the user’s trust in their robot partner.

\section{JOINT ACTION FRAMEWORK}
        \label{sec:system}

\begin{figure}[h]
    \centering
    \includegraphics[trim=0cm 0cm 0cm 0cm, clip, width=0.8\textwidth]{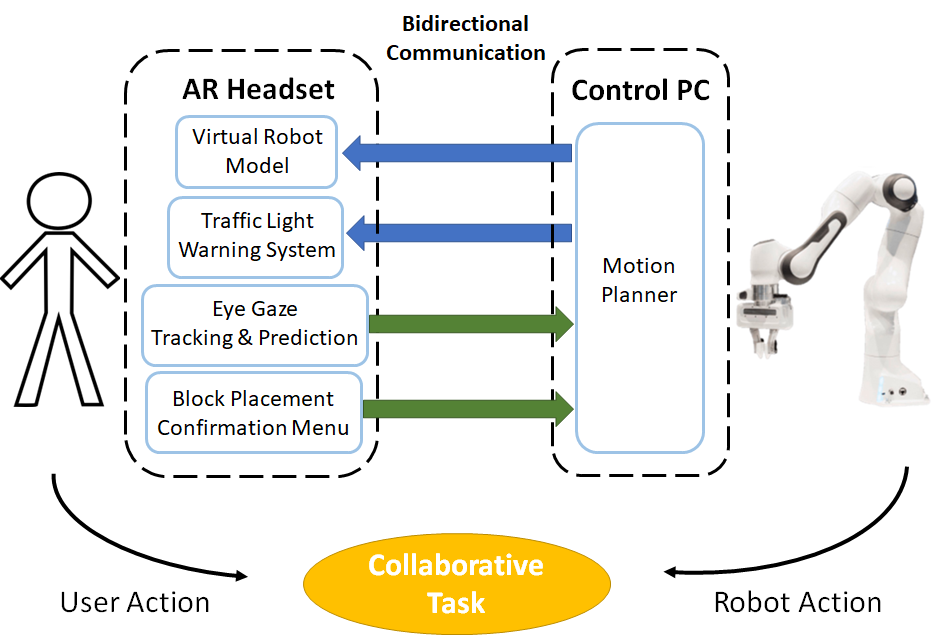}
    \caption{System Diagram}
    \label{fig:systemDiagram}
\end{figure}

We propose a joint action framework implementation that can predict user intent, communicate robot intent, and update the robot’s actions based on its cognitive model of the user (Fig. \ref{fig:systemDiagram}). Our system is comprised of three main components: a) an eye gaze tracking and user intent prediction system, b) a traffic light warning system that conveys the robot's intent to the user through AR displays, and c) a robot motion planner that can update and adjust the robot's actions based on perceived user intent. 

When working on a collaborative task, the user and the robot each formulates an action plan, and executes the actions to contribute towards completing the task. Through the use of AR and eye gaze, our system enables the agents to communicate and predict each other's future actions. Using such information, each agent can then update their cognitive model of the other agent, and adjust their own action plan accordingly.

\begin{figure}[h]
    \centering
    \includegraphics[trim=0cm 0cm 0cm 0cm, clip, width=0.40\textwidth]{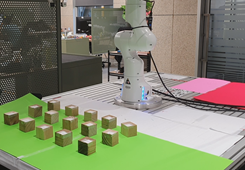}
    \caption{The experiment setup. On the left are the blocks at their initial positions, ready to be picked up by either the robot and the user. Behind the robot are two "block placement zones" which are highlighted with bright coloured (pink and red) papers.}
    \label{fig:expSetup}
\end{figure}

To provide context for the development, implementation, and evaluation of our proposed framework, we consider a collaboration scenario, where a human and a robot needs to move a set of items (blocks) from their starting locations to their respective target locations (Fig. \ref{fig:expSetup}). Such a task is representative of many common tasks in the industry such as packaging, sorting, and assembly.

We implemented the AR interface on a HoloLens 2 head-mounted AR display device\footnote{Microsoft HoloLens 2|https://www.microsoft.com/en-us/hololens}, and the robot motion planning system using Robot Operating System (ROS) on a control PC. We used the Franka Emika Panda robot arm. The following subsections explain the system modules.

\subsection{AR Interface}
\label{AR_Interface}
\textbf{Virtual Robot Model:} We created a one-to-one scale virtual robot model co-located with the real robot. The virtual model is connected to the real robot and mimics the real robot's movements. An AR tag placed at a known location relative to the real robot and the work table is used to calibrate the AR system at startup. The AR headset detects the position of this marker using its outward facing cameras, and it uses its location to compute the relative transformation between the virtual world and the physical world. Once this relative transformation is determined, the virtual robot model and other virtual objects can then be rendered at the appropriate locations. The virtual model allows the user to confirm the calibration and connectivity between the virtual and real robot/system.

\begin{figure}[b]
    \centering
    \includegraphics[trim=0cm 0cm 0cm 0cm, clip, width=0.8\textwidth]{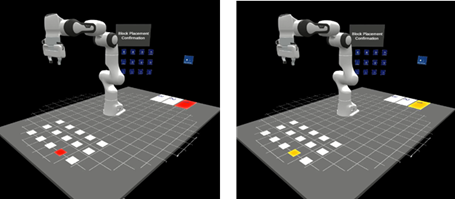}
    \caption{Traffic Light System in action. }
    \label{fig:trafficLightSystem}
\end{figure}

\textbf{Traffic Light Warning System}: 
A traffic light warning system is used to communicate the robot's intent to the user (Fig \ref{fig:trafficLightSystem}). The color red is used to indicate danger and yellow to indicate caution. As this association is well established to most people due to the universal use in traffic lights, users can effortlessly interpret this information. In our system, when the robot selects a target block to pick up and a target zone to drop off, it will highlight both the block it plans to pick and the target placement zone in yellow. This signals to the human that the robot is intending to pick the selected block, and place it in the selected zone. The robot will then move to a ready position, and plan it's pick-and-place motion. After 3 seconds, the system will then highlight the selected block and drop off zone in red to signal to the user that the robot is about to start moving, and the robot will execute the pick-and-place motion.

\begin{figure}[t]
    \centering
    \includegraphics[trim=0cm 0cm 0cm 0cm, clip, width=0.5\textwidth]{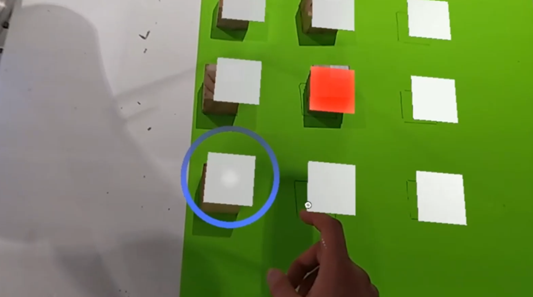}
    \caption{Eye Gaze Tracking and Prediction system in action. The block with blue circle at the bottom are the one going to be picked by the user from predicting their eye gaze.}
    \label{fig:GazeTrackingSystem}
\end{figure}

\begin{figure}[t]
    \centering
    \includegraphics[trim=0cm 0cm 0cm 0cm, clip, width=0.50\textwidth]{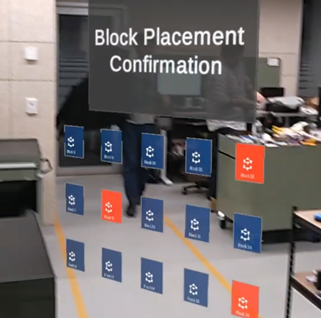}
    \caption{Block Placement Confirmation Menu. Red means the block has been taken by the user. Blue means the block has not been picked up yet, or block that has been picked and placed by the robot.}
    \label{fig:menu}
\end{figure}

\textbf{Eye Gaze Tracking \& Prediction}: 
We use the Hololens' eye tracking function to track the user's gaze and predict their intention (Fig. \ref{fig:GazeTrackingSystem}) since people often gaze at the target object before they manipulate it. The Hololens' eye tracking function can determine where in the workspace the user is gazing at. If the user’s gaze continuously dwells on a block for a certain period of time ($d$ seconds), our system predicts the block as the next target block the user intends to pick and move. Continuous dwell was used to filter out eye saccades that frequently occurs while the user scans the workspace to plan their next moves. For our purposes, the dwell period $d$ was empirically set to 0.8 seconds to balance between prediction speed and accuracy. 

\textbf{Block Placement Confirmation Menu}:
For the purpose of our user study, we implemented an interactive AR window for the user to indicate to the robot which block they have picked (Fig. \ref{fig:menu}). The AR window has a set of buttons with labels corresponding to each of the blocks. The user is asked to press the corresponding AR button each time after they pick up a block. This is implemented as an easy and reliable way to enable to robot to keep track of which blocks are still remaining in the workspace. In a real factory setting, this menu could be be replaced by a camera detection systems.   

\subsection{Robot Motion Planning \& Control System}
The robot motion planner plans the robot actions. It takes into account the predicted user action, and adjusts the robot's action plan accordingly. 
During the collaboration task, at each time, the robot will randomly select its next target block from the remaining blocks on the workspace. It then highlights the block in yellow using the \textbf{Traffic Light Warning System}. While a block is highlighted yellow, if the \textbf{Eye Gaze Tracking \& Prediction} module determines that the user is also intending to pick the same block, the motion planner will decide to pick another block instead. However, once the a target block is highlighted red, the robot is committed to that block, and will proceed to pick the block regardless of what the predicted user target block is. 

After the robot completes the picking action, it will check its gripper finger positions to determine if there is an object between its fingers. If the gripper fingers are fully closed, the robot will infer that the user has taken the block instead. The robot will then abort the placing motion, and plan to pick another block instead. 

The starting positions of the blocks and the placement zone positions are fixed and known to the robot, and we used the MoveIt! Motion Planning Framework for planning the robot's trajectories \cite{moveit}

        
\section{EXPERIMENT}
        \label{sec:experimental_evaluation}
        \subsection{Experimental Design}
We conducted a user study to evaluate the effectiveness of our joint action framework implementation and examine the benefits it provides to human-robot collaboration tasks.
We tested four conditions in our user study, where participants collaborated with the robot using 1) our full framework implementation with both eye gaze prediction and the AR traffic light warning system, 2) only eye-gaze prediction, 3) only the AR traffic light warning system, and 4) and a baseline condition with no eye gaze prediction and no AR displays. A Balanced Latin Square design is used to mitigate ordering effects

The experiment setup is shown in Fig. \ref{fig:expSetup}. There are 15 blocks placed on one side of the workspace, and the human and the robot are required to collaboratively move all the blocks to the other side of the workspace. There are two placement zones labelled ``1" and``2" (red and pink areas in Fig. \ref{fig:expSetup}). Each block also has a label``1" or ``2". Both the participant and the robot can only move one block at a time, and they can work asynchronously. The participant is instructed to place the block they pick into the placement zone with the same label. However, the robot can place each block it picks randomly in either zone. This is to make the robot's action appear not fully predictable to the participant, so that we can test the benefits of our system. Participants were told that when a robot highlights a block/zone yellow, it means that the robot has selected the block/zone. However, the participant may still override the robot's decision by taking the block selected. However, once the block/zone is highlighted red, it means that the robot has committed to it, and is going to start moving and picking up the block, and the participant should not try to pick the highlighted block. 

The rules for the participants are the following: 

\begin{itemize}
    \item The participant cannot enter a zone that the robot has committed to (red zone). 

    \item The participant cannot grab a block after the robot has committed to it.
    
    \item The participant cannot be within 20cm of the robot or in a position that risks causing a collision with the robot.
\end{itemize}

These rules are designed to protect the participant from colliding with the robot. The breaching of any of these rules will result in the E-stop button for the robot being pressed by the experimenter. Both the robot and the participant and the robot are then reset, the blocks they are holding are placed back in their starting position, before the experiment continued. 

The experiment is completed when all the blocks have been moved to the placement zones. After the experiment, participants are asked to complete a collection of surveys, which consists of the NASA Task Load Index \cite{nasa}, 14 items from Trust Perception Scale-HRI \cite{trust} for perception of trust and a subset of questions from the Subjective Fluency Metric scales \cite{fluency}, which are listed in Table \ref{tab:my-table}, for task fluency. We also recorded the time taken to finish the whole task, the number of block picking errors and block placing errors. A block picking error occurs when the participant picks up the block that the robot has committed to pick up. A block placing error occurs when the participant places a block into the same zone the robot is placing its block into.

\begin{table}[t]
\centering
\caption{Subjective Fluency Metric Scales in post experiment survey}
\label{tab:my-table}
\begin{tabular}{p{8cm}p{12cm}}
\hline
Individual Questions \\ \hline
Q1 The human-robot team worked fluently together. \\ 
Q2 The human-robot team's fluency improved over time. \\
Q3 The robot contributed to the fluency of the collaboration. \\ 
Q4 The robot was committed to the success of the team. \\ 
Q5 The robot had an important contribution to the success of the team. \\ \hline

\end{tabular}
\end{table}

\subsection{Hypotheses}
We hypothesize the following:

\textbf{H1}: Our joint action framework implementation using AR and gaze tracking together will increase task efficiency when compared to using AR alone, using Gaze Tracking alone or using neither.

\textbf{H2}: Our joint action framework implementation using AR and gaze tracking will not increase the perceived taskload compared to using AR alone, using Gaze Tracking alone or using neither.

\textbf{H3}: Our joint action framework implementation using AR and gaze tracking together will increase the perception of trust toward the robot compared to using AR alone, using Gaze Tracking alone or using neither.

\textbf{H4}: Our joint action framework implementation using AR and gaze tracking together will increase the fluency of the task compared to using AR alone, using Gaze Tracking alone or using neither.

\section{DATA ANALYSIS \& RESULTS}
        \label{sec:analysis}
        The data were collected from 37 participants (27 males, 7 females, 3 prefer not to say) with a mean age of 22.75 (SD = 2.68). 
Due to COVID-19 restrictions in place at the time of the experiment, only students from Monash University were available to participate in the experiment.
The user study was approved by the Monash University Human Research Ethics Committee (Project ID: 28438). Following existing works \cite{stadler,deWinterDodou,chanTHRI}, we analyzed our data using ANOVA, and post-hoc analysis using t-test and Mann Whitney-Wilcoxon test.
Examining our data using the Anderson-Darling normality test showed that three out of thirty seven sets (8.1\%) were normally distributed. A study by de Winter and Doudou \cite{deWinterDodou} showed that there is only a marginal difference in power between 
using paired t-test and Mann-Whitney-Wilcoxon test when the data is not normally distributed. We analyzed our data with both methods and confirmed that results were marginally different, and paired t-tests will be reported below. Results were treated with Bonferroni's p-value correction \cite{pvalueadjust} and significance reported at $\alpha$ = 0.05.

\subsection{Task Completion Time}

Task completion times are reported in Table \ref{table:time}. The completion times for all conditions were very similar, and ANOVA results indicated that there are no significant differences among all conditions tested for task completion time.

\begin{table}[t]
\caption{Task Completion Time measured in seconds}
\vspace*{-3mm}
\label{table:time}
\centering
\setlength\tabcolsep{2.5pt}
\begin{tabular}{ |c|c|c|c|c| } 
\hline
\textbf  & \textbf{AR and Gaze} & \textbf{AR only} & \textbf{Gaze only} & \textbf{No AR No Gaze} \\
\hline

\textbf{t(s)} & 178.3$\pm$24.4 & 180.7$\pm$24.5 & 180.9$\pm$25.8 & 179.1$\pm$29.6 \\  

\hline
\end{tabular}
\vspace*{-1mm}
\end{table}

\begin{table}[t]
\caption{Number of Block Picking and Placing errors in each trial}
\vspace*{-3mm}
\label{table:error}
\centering
\setlength\tabcolsep{1.5pt}
\begin{tabular}{ |c|c|c|c|c| } 
\hline
\textbf  & \textbf{AR and Gaze} & \textbf{AR only} & \textbf{Gaze only} & \textbf{No AR No Gaze} \\
\hline

\textbf{Placing Error} & 0.19$\pm$0.39 & 0.37$\pm$0.54 & 0.65$\pm$0.81 & 1.27$\pm$0.95 \\ 
\textbf{Picking Error} & 0.05$\pm$0.23 & 0.16$\pm$0.49 & 0.38$\pm$0.54 & 0.49$\pm$1.54 \\  

\hline
\end{tabular}
\vspace*{-1mm}
\end{table}

\subsection{Block Placing and Picking Errors}
Number of block placing and picking errors are reported in Table \ref{table:error}. ANOVA results indicated that there are significant differences in block placing errors among the four conditions (F(3, 144) = 16, p $<$ 0.001), while there are no differences in block picking errors among the four conditions. Post-hoc analysis showed that our joint action framework (AR and Gaze Tracking) significantly reduced block placing errors when compared to No AR no Gaze, i.e. baseline, (t(36) = -6.8938, p $<$ 0.001) and when compared to Gaze Tracking only (t(36) = -3.0026, p = 0.0145). 

\begin{table}
\caption{NASA-TLX \cite{nasa} questionnaire results (7 point scale). For Performance, high score is better; for the rest, lower score is better}
\vspace*{-3mm}
\label{table:TLX}
\centering
\setlength\tabcolsep{1.5pt}
\begin{tabular}{ |c|c|c|c|c| } 
\hline
\textbf  & \textbf{AR and Gaze} & \textbf{AR only} & \textbf{Gaze only} & \textbf{No AR No Gaze} \\
\hline

\textbf{Mental} & 2.08$\pm$1.07 & 2.30$\pm$1.27 & 2.81$\pm$1.54 & 2.65$\pm$1.60 \\  
\textbf{Physical} & 1.92$\pm$1.0 & 1.84$\pm$0.85 & 2.05$\pm$1.21 & 1.95$\pm$1.18 \\  
\textbf{Temporal} & 2.38$\pm$1.26 & 2.46$\pm$1.39 & 2.43$\pm$1.46 & 2.73$\pm$1.5 \\  
\textbf{Performance} & 1.97$\pm$1.48 & 2.35$\pm$1.91 & 2.16$\pm$1.24 & 2.22$\pm$1.49 \\  
\textbf{Effort} & 2.03$\pm$1.03 & 2.43$\pm$1.50 & 2.62$\pm$1.58 & 2.57$\pm$1.50 \\  
\textbf{Frustration} & 2.13$\pm$1.36 & 2.00$\pm$1.27 & 2.49$\pm$1.69 & 2.35$\pm$1.65 \\  

\hline
\end{tabular}
\vspace*{-1mm}
\end{table}

\subsection{Perceived Taskload}
Results indicated that there were no significant differences among the four conditions for perceived taskload measured by the NASA-TLX.
Results of all four conditions are summarized in Table \ref{table:TLX}.

\begin{table}
\caption{Perception of Trust \cite{trust} survey results (10 point scale). Asterisks indicate significant differences found.}
\vspace*{-2mm}
\label{table:trust}
\centering
\setlength\tabcolsep{1.5pt}
\begin{tabular}{ |c|c|c|c|c| } 
\hline
\textbf  & \textbf{AR \& Gaze} & \textbf{AR only} & \textbf{Gaze only} & \textbf{No AR/Gaze} \\
\hline

\textbf{Dependable} & 7.92$\pm$2.10 & 7.11$\pm$2.73 & 7.08$\pm$2.61 & 6.81$\pm$2.26         \\
\textbf{Reliable} & 8.19$\pm$1.83 & 6.81$\pm$2.80 & 7.08$\pm$2.57 & 7.24$\pm$2.12 \\
\textbf{Unresponsive*} & 1.81$\pm$2.43 & 1.51$\pm$1.80 & 2.84$\pm$3.10 & 2.97$\pm$3.10 \\
\textbf{Predictable*} & 7.16$\pm$2.85 & 6.38$\pm$3.22 & 5.22$\pm$3.11 & 4.73$\pm$3.05 \\
\textbf{Consistent*} & 8.08$\pm$2.19 & 7.43$\pm$2.76 & 6.51$\pm$2.77 & 7.00$\pm$2.53 \\
\textbf{Successful} & 8.30$\pm$1.71 & 8.24$\pm$1.95 & 7.59$\pm$2.33 & 8.27$\pm$1.67 \\
\textbf{Malfunction} & 0.97$\pm$1.85 & 1.97$\pm$2.90 & 1.89$\pm$2.67 & 1.76$\pm$2.69 \\
\textbf{Have Error} & 1.41$\pm$2.03 & 1.78$\pm$2.21 & 1.89$\pm$2.24 & 1.51$\pm$2.10 \\
\textbf{Provide Feedback*} & 4.54$\pm$4.32 & 5.68$\pm$4.05 & 1.73$\pm$2.85 & 1.49$\pm$2.96 \\
\textbf{Meet Needs} & 7.65$\pm$2.77 & 7.70$\pm$2.38 & 7.54$\pm$2.34 & 7.57$\pm$2.27 \\
\textbf{Informative*} & 8.00$\pm$2.49 & 7.38$\pm$3.10 & 2.95$\pm$3.73 & 2.54$\pm$3.62 \\
\textbf{Communicative*} & 6.95$\pm$3.46 & 6.62$\pm$3.46 & 2.03$\pm$3.12 & 1.54$\pm$2.86 \\
\textbf{Do as Instructed} & 8.30$\pm$2.29 & 8.05$\pm$2.36 & 7.03$\pm$3.17 & 7.70$\pm$2.43 \\
\textbf{Follow Directions} & 7.86$\pm$2.79 & 7.35$\pm$3.11 & 6.11$\pm$3.70 & 6.89$\pm$3.34 \\

\hline
\end{tabular}
\vspace*{-1mm}
\end{table}

\subsection{Trust Perception}
Survey results of trust perception are shown in Table \ref{table:trust}. ANOVA results showed that there were significant differences for several items - Unresponsive: (F(3, 144) = 2.71, p = 0.0472), Predictable: (F(3, 144) = 4.68, p = 0.0038, Provide Feedback: (F(3, 144) = 11.96, p $<$ 0.001), Informative: (F(3, 144) = 27.7. p $<$ 0.001), Communicative: (F(3, 144) = 28.83, p $<$ 0.001)) and fluency (Fluently Improved: (F(3, 144) = 3.22, p = 0.025).

Post-hoc analysis revealed that our system combining AR and Gaze Tracking improved user’s trust toward the robot in some aspects compared to the baseline. Our system received better ratings for the following items: Predictable (t(36) = 4.0142, p $<$ 0.001), Provide Feedback (t(36) = 3.3763, p = 0.0053), Informative (t(36) = 6.8456, p $<$ 0.001) and Communicative (t(36) = 6.6544, p $<$ 0.001). Our system when compared to the Gaze-only system received better ratings for the items Consistent (t(36) = 2.5558, p = 0.0449), Provide Feedback (t(36) = 4.8061, p $<$ 0.001), Informative (t(36) = 7.095, p $<$ 0.001) and Communicative (t(36) = 5.715, p $<$ 0.001). The combined system when compared to AR-only system did not have any significant differences.

\begin{table}
\caption{Subjective Fluency Metrics \cite{fluency} results (7 point scale). Higher score is better}
\label{table:fluency}
\centering
\setlength\tabcolsep{1.5pt}
\begin{tabular}{ |c|c|c|c|c| }
\hline
\textbf  & \textbf{AR \& Gaze} & \textbf{AR only} & \textbf{Gaze only} & \textbf{No AR/Gaze} \\
\hline

\textbf{Fluently Together} & 5.73$\pm$1.31 & 5.43$\pm$1.33 & 4.67$\pm$1.8 & 5.05$\pm$1.63 \\ \hline 
\textbf{Fluency Improved} & 5.41$\pm$1.55 & 5.43$\pm$1.62 & 4.92$\pm$1.75 & 5.20$\pm$1.48 \\  \hline 
\textbf{Robot Fluency} & 5.62$\pm$1.60 & 5.21$\pm$1.74 & 4.65$\pm$1.98 & 4.78$\pm$1.74 \\  
\textbf{Contribute} & & & & \\  \hline 
\textbf{Robot Committed} & 5.41$\pm$1.75 & 5.05$\pm$1.86 & 4.78$\pm$1.9 & 4.95$\pm$1.74 \\  \hline 
\textbf{Robot Success} & 5.51$\pm$1.67 & 5.49$\pm$1.69 & 5.08$\pm$1.85 & 5.03$\pm$1.65 \\  
\textbf{Contribution} & & & & \\ \hline 

\hline
\end{tabular}
\vspace*{-1mm}
\end{table}

\subsection{Fluency}
Table \ref{table:fluency} summarizes the results for the survey questions on fluency. Our system combining the use of AR and Gaze yielded the best scores for all questions, except for fluency improvements, which had higher score only in the AR condition. ANOVA results showed that there were significant differences in Q1 of the fluency survey - Fluently together (F(3, 144) = 3.2238, p = 0.0245).
Post-hoc analysis indicated that with the help of the AR and Gaze Tracking system combined, users felt that the human-robot team worked together more fluently (Q1) compared to using Gaze Tracking-only system (t(36) = 2.8368, p = 0.0223). Other than that, there were no statistically significant differences.

\section{DISCUSSION}
        \label{sec:discussion}
        \subsection{Task Efficiency}
Our hypothesis \textbf{H1} stated that our joint action framework implementation using AR
and gaze tracking will increase task efficiency, and our results supported this claim. 
Experiment results showed that our system combining AR and Gaze Tracking yielded fewer block placing errors when compared to the baseline or the Gaze Tracking only system.
While in our experiment, block placing errors did not result in any penalties or other consequences, in certain human-robot collaborative tasks, making errors could potentially result in a costly recovery procedure, or even injuring a human. In such cases, the benefits provided by our joint action framework by reducing task errors would be even greater.

\subsection{Perceived Taskload}
Results shows that our system combining AR and Gaze tracking did not increase the perceived taskload, and this provided support to \textbf{H2}. It is an advantage that our system does not come with the cost of an increased perceived taskload, as some existing studies have found that the use of AR can result in increased mental demand \cite{quintero_robot_2018}. There are a couple of differences between our study the the one in  \cite{quintero_robot_2018}. 
Our study involves a human-robot collaboration task, and our system uses a traffic light warning system, which is highly familiar to most users, for AR display. The study in  \cite{quintero_robot_2018}, on the other hand, involved a robot trajectory programming task, and uses curves and spherical markers to display robot trajectory, which is less familiar to users, in AR. As AR is still a relatively novel technology to users, it may be important to choose more familiar concepts for implementing AR visual displays when possible, to avoid increasing the users' cognitive load.


\subsection{Perception of Trust}
Our results have provided support to \textbf{H3}, which stated that our joint action framework implementation using AR and gaze tracking together will increase the perception of trust. As reported in the previous section, our system yield better scores in almost all items in the trust perception survey, when compared to all other conditions. In a team collaboration setting, bidirectional communication and establishing mutual understanding is important towards building trust. Our joint action framework system aims to enable to establishment of mutual understanding by enabling bidirectional communication between collaborating humans and robots. Through our study, we have demonstrated that our system enables better establishment of trust between human and robot partners.

\subsection{Task Fluency}
Our results showed that our system yielded best scores for almost all task fluency survey questions, although not reaching significance. Thus we found weak support for \textbf{H4}. We suspect that this may be a result of the limited accuracy of the gaze tracking system of the Hololens, as we found that the gaze prediction system only had an accuracy of around 50\% when predicting which block the user actually picks in the experiment. Improving the gaze tracking system's performance can potentially further improve the task fluency when using our overall system, and this can be a direction of future work.

\section{CONCLUSION}
        \label{sec:conclusion}
        In this paper, we have proposed the design of a novel joint action framework for facilitating efficient human-robot collaboration. Our system enables bidirectional communication via the use of AR to communicate robot intent, and eye gaze to predict user’s task intent. We have presented an implementation of the framework and a user study to demonstrate the benefits of our proposed system. Results showed that our system increases task efficiency, trust, as well as task fluency, while not imposing a higher perceived taskload, which is a critical outcome for creating more functional human-robot collaborative systems without burdening the user. Potential future work may focus on improving the performance of gaze tracking, as we believe this can further improve the performance of our joint action framework system, or further applying and testing the benefits of our system for highly asynchronous tasks.

\begin{acks}
\label{sec:acknowledgement}
This project was supported by the Australian Research Council Discovery
Projects Grant, Project ID: DP200102858.
\end{acks}

\bibliographystyle{ACM-Reference-Format}
\bibliography{ref}

\end{document}